\documentclass[journal]{IEEEtran}

\usepackage{cite}           
\usepackage{hyperref}       
\usepackage{url}            
\usepackage{booktabs}       
\usepackage{amsfonts}       
\usepackage{microtype}      
\usepackage{amsmath}
\usepackage{graphicx}       
\graphicspath{{figures/}}
\hypersetup{
  colorlinks=true,
  linkcolor=blue,
  citecolor=blue,
  urlcolor=blue
}

\renewcommand{\eqref}[1]{\textup{\begingroup\color{blue}(\ref{#1})\endgroup}}

\title{Dual-Domain Self-Supervised Artifact Removal Framework for Photoacoustic Computed Tomography}

\author{Yucheng~Zhou, Shuang~Li, Yu~Zhang, Yibing~Wang, Chulhong~Kim,
Seongwook~Choi, and~Changhui~Li%
\thanks{Y. Zhou, S. Li, Y. Zhang, Y. Wang, and C. Li are with the Department of Biomedical Engineering, College of Future Technology, Peking University, Beijing, China (e-mail: chli@pku.edu.cn).}%
\thanks{C. Kim and S. Choi are with the Departments of Electrical Engineering, Convergence IT Engineering, Mechanical Engineering, and Medical Science and Engineering, and the Medical Device Innovation Center, Pohang University of Science and Technology, Pohang, Republic of Korea.}%
\thanks{Corresponding author: Changhui Li.}}

\begin{document}
\maketitle

\begin{abstract}
Photoacoustic Computed Tomography (PACT) often faces severe challenges from reconstruction artifacts due to sparse detection conditions. In this work, based on the distinct differences in artifact patterns between back-projection-based and Fourier-based reconstruction algorithms, we propose a self-supervised artifact removal framework that employs a lightweight Siamese Neural Network and a composite loss function integrating cross-domain fidelity and uncertainty-weighted consistency, effectively decoupling dual-domain features and filtering artifacts. Comprehensive validations using simulations, phantoms, in vivo rat and human experimental data demonstrate that the proposed method can significantly suppress image artifacts. Furthermore, enabled by the acceleration of the spatial-domain and frequency-domain inverse operator, this end-to-end approach also achieves exceptional computational efficiency.
\end{abstract}

\begin{IEEEkeywords}
Photoacoustic computed tomography, artifact removal, unsupervised learning, dual-domain learning.
\end{IEEEkeywords}

\section{Introduction}
Photoacoustic computed tomography (PACT), uniquely combining the advantages of optical contrast with deep penetration and high resolution of ultrasound, has emerged as a highly promising non-invasive imaging modality in biomedicine\cite{Li_2009}\cite{park2025clinical}\cite{lin2022emerging}\cite{ntziachristos2025addressing}. PACT has been successfully applied to the non-invasive investigation of vascular imaging\cite{huynh2025fast}\cite{10.1117/1.JBO.29.S1.S11519}\cite{ivankovic2019real}, tumor detection\cite{wang2016practical}\cite{lin2022emerging}, and brain functions\cite{wang2003noninvasive}\cite{na2022massively}\cite{chen2024photoacoustic}. Constrained by the geometry of the imaged target, hardware costs and system complexity, practical PACT systems generally have a limited number of sensor elements or limited-view coverage\cite{tian2021spatial}. Consequently, traditional back-projection(BP) algorithms, such as Delay-and-Sum (DAS) and Universal Back-Projection (UBP)\cite{xu2005universal}, as well as Fourier-based algorithms\cite{kunyansky2011fast}\cite{treeby2010k}, all suffer from severe structural image artifacts.

While traditional image denoising methods (e.g., BM4D\cite{6253256} and Noise2Noise\cite{lehtinen2018noise2noiselearningimagerestoration}) can mitigate statistical noise in PACT, they generally fail to eliminate the complex structural artifacts caused by sparse sampling. To address this issue with physical constraints, model-based iterative reconstruction (MBIR) methods have been widely explored\cite{10.3788/PI.2024.R06}\cite{6749001}\cite{10111086}. Recent advancements have successfully extended MBIR to full 3D reconstruction\cite{7927480} and significantly improved its memory efficiency\cite{li2025slingbag}. However, MBIR inherently requires repeated forward and backward physical projections, which demand substantial computational resources.

In recent years, deep learning-based methods have provided novel paradigms for artifact removal.\cite{davoudi2019deep}\cite{https://doi.org/10.1002/advs.202202089}\cite{grohl2021deep}\cite{kim2020deep}\cite{rajendran2022photoacoustic}\cite{jeong2026hybrid} However, conventional supervised learning networks heavily rely on large-scale, high-quality paired data for training, which is difficult in practice. Therefore, unsupervised algorithms have gained much attention. The blind-spot strategy achieves effective self-supervised artifact removal by intentionally masking a portion of the input data \cite{LI2025100723}\cite{zhang2025iterative}\cite{lan2024masked}. However, since sparse reconstruction is inherently a highly ill-posed inverse problem, introducing a secondary mask inevitably exacerbates data scarcity and compromises the physical integrity of the acoustic measurements, thereby influencing the performance of artifact-removal performance. Alternatively, the Deep Image Prior (DIP) method~\cite{ulyanov2018deep} relies on network architecture as an implicit prior to enhance the quality of  photoacoustic microscopy (PAM) images\cite{zhang2023adaptive}\cite{vu2021deep}. However, because DIP is fitted to each corrupted measurement without external training data or explicit artifact modeling, the network may gradually learn not only the underlying anatomical structures but also the deterministic and spatially correlated artifacts present in PACT. This tendency makes it difficult to separate structured artifacts from true anatomical features. In addition, generative models based on unpaired datasets have also been explored for photoacoustic reconstruction\cite{paul2024enhancement}\cite{lu2021artifact}. Yet, these methods rely heavily on statistical priors to compensate for missing physical information, which frequently cause hallucinate fake vascular networks

This study reports a self-supervised artifact removal method that leverages the distinct artifact behaviors arising from the spatial-domain UBP and the frequency-domain based method (FDM). Specifically, UBP artifacts arise from sparse spatial sampling, whereas frequency-domain artifacts stem from spectral truncation and aliasing\cite{hu2020spatiotemporal}. We then propose a self-supervised complementary information learning framework. This framework employs a lightweight, weight-sharing convolutional neural network to estimate artifact residuals from the reconstructed dual-domain reconstructions. By introducing a joint constraint comprising cross-domain fidelity loss and consistency loss, the network is compelled to effectively learn and decouple different artifact representations from the multi-source inputs. Consequently, this enables effective self-supervised artifact removal directly in the post-reconstruction stage.

\section{Method}
\begin{figure*}[!t]
    \centering
    \includegraphics[width=\textwidth]{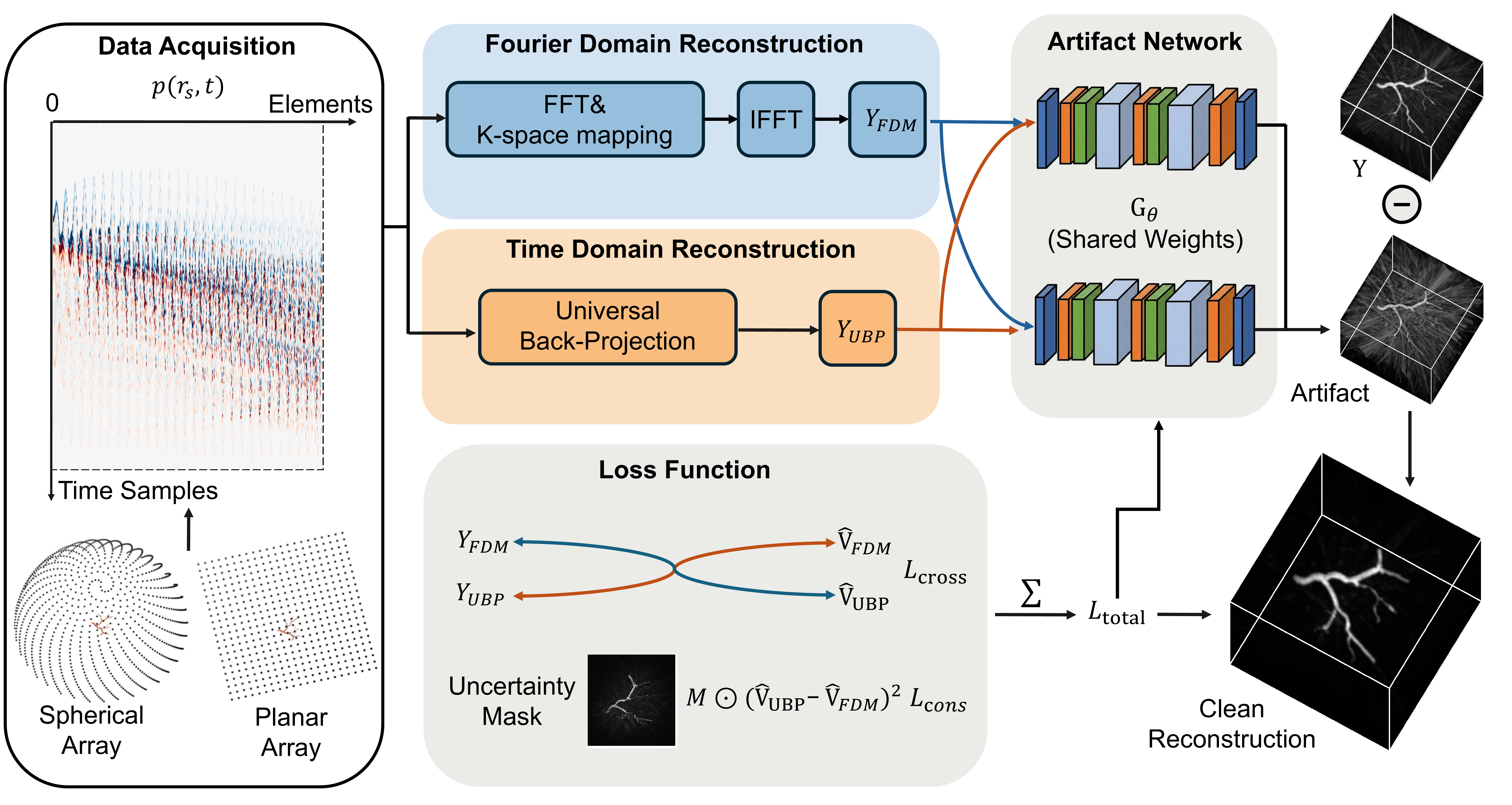}
    \caption{Dual-domain self-supervised artifact removal reconstruction framework. The workflow includes sparse sampling data acquisition, two initial reconstructions via FDM and UBP, a Siamese artifact network, and a physics-informed joint loss module.}
    \label{fig1}
\end{figure*}

This section details the signal forward modeling, the geometry-specific fast Fourier-domain inversion operators, the universal spatial baseline algorithm, and the proposed artifact removal network based on complementary information learning. An overview of the proposed dual-domain self-supervised reconstruction framework is shown in Fig.~\ref{fig1}.

This study assumes a homogeneous, non-viscous acoustic medium. Let $\Omega$ denote the detection surface, $p_0(\mathbf{r})$ denote the initial PA pressure, and $p(\mathbf{r}_s, t)$ denote the time-domain pressure signal received by a transducer located at $\mathbf{r}_s$ on $\Omega$.
\subsection{Detection Geometry}

Linear/Planar Array: Transducers are arranged along a line in  two-dimensional (2D) or in a 2D matrix on a plane in  three-dimensional (3D), typically the $z=0$ plane.

Circular/Hemispherical Array: Transducers are distributed along a circle or arc in 2D, or on a hemispherical surface with radius $R_s$ in 3D.

\subsection{Image Reconstruction Algorithms}

\subsubsection{Universal Back-Projection (UBP) Reconstruction}
The spatial-domain UBP reconstruction provides a direct inversion, as follows\cite{xu2005universal}:
\begin{equation}
\label{eq:ubp}
    p_{\mathrm{UBP}}(\mathbf{r}) = \int_{\Omega} d\Omega \left[ 2p(\mathbf{r}_s, t) - 2t \frac{\partial p(\mathbf{r}_s, t)}{\partial t} \right]_{t = |\mathbf{r} - \mathbf{r}_s|/v_s}
\end{equation}
where $p_{\mathrm{UBP}}(\mathbf{r})$ represents the reconstructed initial PA pressure at spatial position $\mathbf{r}$. The integration is performed over the detection surface $\Omega$, where $p(\mathbf{r}_s, t)$ is the time-domain pressure signal detected by the sensor at $\mathbf{r}_s$. The term $t$ represents the time-of-flight (ToF), determined by the Euclidean distance $|\mathbf{r} - \mathbf{r}_s|$ and the speed of sound $v_s$. 
\subsubsection{Fourier-Domain (FDM) Reconstruction}\mbox{}\par
\noindent\textbf{Circular and spherical geometries:} following the Fourier-domain reconstruction proposed by Kunyansky\cite{kunyansky2011fast}, the measured PA signals are first converted into the Fourier transform of the initial pressure distribution in $k$-space, after which the initial pressure distribution is reconstructed using an inverse FFT, as follows:
\begin{equation}
\label{eq:pfft_recon}
p_{\mathrm{FDM}}(\mathbf r) 
=
\mathcal F_d^{-1}
\left\{
\mathcal G\left[\widetilde p(\mathbf k)\right]
\right\}(\mathbf r),
\qquad
k=|\mathbf k|=\omega/v_s .
\end{equation}
Here, $p_{\mathrm{FDM}}(\mathbf r)$ denotes the reconstructed initial PA pressure at the spatial position $\mathbf r$. The parameter $d$ denotes the reconstruction dimension, with $d=2$ for a 2D circular detection array and $d=3$ for a 3D spherical detection array. The $\widetilde p(\mathbf k)$ represents the spectrum of the initial pressure distribution in the wavenumber domain, or $k$-space, where $\mathbf k$ is the wavenumber vector and $k=|\mathbf k|$ is its magnitude. The variable $\omega$ denotes the angular frequency, and $v_s$ is the speed of sound in the medium; the operator $\mathcal G[\cdot]$ denotes density-compensated gridding, which interpolates the non-Cartesian $k$-space samples $\widetilde p(\mathbf k)$ acquired on a polar or spherical grid onto a Cartesian grid. The operator $\mathcal F_d^{-1}$ denotes the $d$-dimensional inverse Fourier transform, which is numerically implemented using a $d$-dimensional inverse fast Fourier transform (IFFT).

For a 2D circular array of radius $R_s$, the spectrum of the initial pressure distribution in the wavenumber domain, $\widetilde p(\mathbf k)$, can be written in polar coordinates as $\widetilde p(k,\phi_k)$\cite{kunyansky2011fast}:
\begin{equation}
\label{eq:pfft_2d}
\widetilde p(k,\phi_k)
=
\sum_{m=-m_{\max}}^{m_{\max}}
\frac{
2(-i)^m \, \mathcal W_m(k) \, P_m(k)
}{
\pi k H_{|m|}^{(1)}(kR_s)
}
e^{im\phi_k},
\end{equation}
where $\phi_k$ is the azimuthal angle of $\mathbf k$ in the 2D wavenumber space. The quantity $\widetilde p(k,\phi_k)$ denotes the Fourier spectrum of the initial pressure distribution in 2D $k$-space. The factor $e^{im\phi_k}$ is the angular Fourier basis
function describing the angular dependence in wavenumber space. The index $m$ denotes the azimuthal Fourier mode, and $m_{\max}$ is the truncation order of the angular expansion, corresponding to the highest angular spatial-frequency component retained in the circular-array reconstruction. $P_m(k)$ is the $m$-th angular Fourier coefficient of the measured PA pressure onto the circular array. $H_{|m|}^{(1)}(\cdot)$ is the cylindrical Hankel function of the first kind and order $|m|$, which represents the radial component of the outgoing acoustic Green's function in the 2D circular geometry. The function $\mathcal W_m(k)$ is a bounded spectral regularization window. In the ideal noise-free formulation, $\mathcal W_m(k)=1$; in practical implementations, it is used to suppress noise amplification and numerical instability caused by small values of $H_{|m|}^{(1)}(kR_s)$ or by ill-conditioned high-order angular modes.

For a 3D spherical array of radius $R_s$, the spectrum of the initial pressure distribution in the wavenumber domain, $\widetilde p(\mathbf k)$, can be written in spherical coordinates as $\widetilde p(k,\theta_k,\phi_k)$\cite{kunyansky2011fast}:
\begin{equation}
\label{eq:pfft_3d}
\widetilde p(k,\theta_k,\phi_k)
=
\sum_{l=0}^{l_{\max}}
\sum_{m=-l}^{l}
\frac{
\sqrt{2/\pi}\, i^l \, \mathcal W_l(k) \, P_{lm}(k)
}{
k^2 h_l^{(1)}(kR_s)
}
Y_l^m(\theta_k,\phi_k),
\end{equation}
where $\mathbf k$ is the 3D wavenumber vector, $k=|\mathbf k|$ is its magnitude, and $\theta_k$ and $\phi_k$ are the polar and azimuthal angles of $\mathbf k$ in 3D wavenumber space, respectively. The quantity $\widetilde p(k,\theta_k,\phi_k)$ denotes the Fourier spectrum of the initial pressure distribution in 3D $k$-space. $Y_l^m(\theta_k,\phi_k)$ is the spherical harmonic of degree $l$ and order $m$, describing the angular dependence in wavenumber space. The index $l$ denotes the spherical-harmonic degree, while $m$ denotes the azimuthal order associated with each degree, with $-l\le m\le l$. The parameter $l_{\max}$ is the truncation order of the spherical-harmonic expansion, corresponding to the highest angular spatial-frequency component retained in the spherical-array reconstruction. $P_{lm}(k)$ is the spherical-harmonic coefficient of the measured acoustic pressure on the detection sphere, obtained from the frequency-domain pressure data measured by the spherical detection array. $h_l^{(1)}(\cdot)$ is the spherical Hankel function of the first kind and degree $l$, which represents the radial component of the outgoing acoustic Green's function in the 3D spherical geometry. Similarly, The function $\mathcal W_l(k)$ is a bounded spectral regularization window. 

In Eqs. \eqref{eq:pfft_2d} and \eqref{eq:pfft_3d}, the angular expansion coefficients $P_\nu(k)$ required to compute the Fourier spectrum of the initial pressure distribution in $k$-space, denoted as $P_m(k)$ for a 2D circular array and $P_{lm}(k)$ for a 3D spherical array, are determined by performing an angular expansion of the frequency-domain pressure $P(\mathbf{r}_s, k)$, i.e., the temporal Fourier transform of the pressure $p(\mathbf{r}_s, t)$ measured on the detection surface:
\begin{equation}
\label{eq:fft}
\begin{aligned}
P(\mathbf{r}_s,k)
  &= \sum_{\nu} P_{\nu}(k)\Phi_{\nu}(\mathbf{r}_s,k), \\
\Phi_{\nu}(\mathbf{r}_s,k)
  &= \mathcal{H}_{\nu}^{(1)}(kR_s)\Psi_{\nu}(\Omega_s).
\end{aligned}
\end{equation}
where $\mathbf{r}_s = (R_s, \Omega_s)$ denotes the spatial coordinates on the boundary of the detection surface, $R_s$ is the geometric radius of the detection array, and $\Omega_s$ represents the corresponding angular coordinates.Here, $P_\nu(k)$ is its expansion coefficient for the $\nu$-th angular mode, and $\Phi_\nu$ is the corresponding basis function.
For spherical arrays, $\nu=(l,m)$, $\mathcal H_{\nu}^{(1)}=h_l^{(1)}$, and $\Psi_{\nu}=Y_l^m(\theta_s,\phi_s)$. For circular arrays, $\nu=m$, $\mathcal H_{\nu}^{(1)}=H_{|m|}^{(1)}$, and $\Psi_{\nu}=e^{im\phi_s}$.
The zero-frequency component is obtained from the finite low-frequency limit
and inserted into the Cartesian $k$-space grid before the final inverse FFT.

\textbf{Planar geometry:}
We implement Stolt migration \cite{treeby2010k} to analytically reconstruct the spatial wavefield. Data are mapped from the frequency domain into wavenumber space governed by the dispersion relation $k_z = \sqrt{(\omega/v_s)^2 - k_{\parallel}^2}$, where $k_{\parallel}$ and $k_z$ denote the transverse and axial wavenumber components, respectively. To rigorously preserve energy during this coordinate transformation, a Jacobian scaling factor is applied, followed by a multidimensional IFFT to yield the final reconstructed image.

\subsection{Mechanism of Reconstruction Artifact}

To explore different mechanisms of reconstruction artifacts in the spatial and frequency domains, the sparse sampling process is modeled as the product of the continuous spatial signal and a Dirac comb function $S(\mathbf{r}_d)$:
\begin{equation}
S(\mathbf{r}_d) = \sum_{i=1}^{N} \delta(\mathbf{r}_d - \mathbf{r}_i)
\end{equation}
where $\mathbf{r}_d$ denotes the coordinates on the detection surface, $\delta$ is the Dirac delta function, and $\mathbf{r}_i$ represents the position of the $i$-th sensor. In PACT, to satisfy the spatial Nyquist criterion, the local spatial sampling interval on the detection surface must be no larger than half of the shortest detectable acoustic wavelength\cite{hu2020spatiotemporal}; otherwise, sparse sampling induces unavoidable aliasing artifacts in the reconstructed image.

\subsubsection{UBP Artifacts}
In PACT,  as Eq. (\ref{eq:ubp}), continuous spatial sampling theoretically yields an exact reconstruction of the initial pressure via the continuous solid-angle integration of the back-projection term,. In sparse-array configurations, this continuous integral structurally degenerates into a sparsely discrete summation over a finite number of detectors, in which the complete destructive interference of back-projected waves outsides real targets breaks down.
Mathematically, the reconstruction $\hat{p}_{\mathrm{UBP}}(\mathbf{r})$ is governed by the superposition of the back-projection term $b(\mathbf{r}_i, t)=2p(\mathbf{r}_i, t) - 2t \frac{\partial p(\mathbf{r}_i, t)}{\partial t}$, as in the following formula:

\begin{equation}
\label{ubpa}
\hat{p}_{\mathrm{UBP}}(\mathbf{r}) = \sum_{i=1}^{N} \Delta \Omega_i b\left(\mathbf{r}_i, \frac{|\mathbf{r} - \mathbf{r}_i|}{v_s}\right) 
\end{equation}

where $\Delta \Omega_i$ denotes the discrete solid-angle weighting for the $i$-th sensor. Eq. (\ref{ubpa}) indicates that each detector back-projects its filtered signal $b(\mathbf{r}_i, t)$ onto an arc (in 2D) or a spherical surface (in 3D) centered on itself. Because spatial undersampling precludes the exact phase cancellation outside of the true acoustic source,  this residual error spatially accumulates, manifesting as highly directional, streak-like aliasing artifacts along specific geometric projection trajectories.
\subsubsection{FDM Artifacts}

Artifacts in Fourier-domain reconstruction can be understood through the modulation of the boundary pressure spectrum by discrete detector sampling. Let $\mathbf{r}_d$ denote the detector coordinate on the measurement surface $\Omega$, and let $P(\mathbf{r}_d,k)$ denote the frequency-domain acoustic pressure defined above, namely, the temporal Fourier transform of the measured signal $p(\mathbf{r}_d,t)$. At a fixed wavenumber $k=\omega/v_s$, the spatial sampling function of a finite detector array is defined as
\begin{equation}
    \ S(\mathbf{r}_d)=\sum_{i=1}^{N}w_i\delta(\mathbf{r}_d-\mathbf{r}_i)\
\end{equation}

where $\mathbf{r}_i$ and $w_i$ denote the location and quadrature weight of the $i$th detector, respectively. The sparsely sampled frequency-domain pressure is therefore given by
\begin{equation}
    \ P_s(\mathbf{r}_d,k) = P(\mathbf{r}_d,k)S(\mathbf{r}_d)\
\end{equation}
Let $\mathcal{F}_{\Omega}$ denote the spatial Fourier transform with respect to the detector coordinate $\mathbf{r}_d$. The spatial spectrum of the sampling function is then
\begin{equation}
    \mathcal{F}_S(\boldsymbol{\kappa})=\mathcal{F}_{\Omega}\left\{S(\mathbf{r}_d)\right\}=\sum_{i=1}^{N}w_i\exp\left(-\mathrm{i}\boldsymbol{\kappa}\cdot\mathbf{r}_i\right)\
\end{equation}

where $\boldsymbol{\kappa}$ is the spatial-frequency variable conjugate to $\mathbf{r}_d$. According to the Fourier product theorem, discrete sampling on the measurement surface corresponds to a convolution between the continuous pressure spectrum and the sampling spectrum in the spatial-frequency domain:
\begin{equation}
    \
\widetilde{P}_s(\boldsymbol{\kappa},k)
=
\frac{1}{(2\pi)^{d_{\Omega}}}
\left[
\widetilde{P}(\cdot,k)
*
F_S
\right]
(\boldsymbol{\kappa}),
\
\end{equation}

where $\widetilde{P}(\boldsymbol{\kappa},k)=\mathcal{F}_{\Omega}\{P(\mathbf{r}_d,k)\}$ denotes the spatial spectrum of the continuous frequency-domain pressure, $d_{\Omega}$ is the intrinsic dimension of the measurement surface, and $*$ denotes convolution. The normalization factor depends on the adopted Fourier-transform convention.

For a uniformly sampled sparse array, the sampling spectrum consists of periodic components in reciprocal space, with their spacing determined by the detector pitch. When the detector pitch violates the spatial Nyquist criterion, adjacent spectral replicas overlap. Consequently, high-spatial-frequency components are folded into the low-frequency region, resulting in irreversible spectral aliasing. For a practical finite array, the finite aperture and nonuniform detector distribution further broaden the spectral replicas and perturb their strict periodicity.

During FDM reconstruction, these aliased components are mapped to the object $k$-space through geometry-dependent modal inversion or Stolt interpolation and subsequently spread throughout the reconstruction field of view after gridding and inverse Fourier transformation. Consequently, FDM artifacts typically appear as globally distributed oscillatory sidelobes or quasi-periodic interference patterns. Their spatial characteristics are jointly determined by the detector pitch, array geometry, finite aperture, and acoustic bandwidth of the imaging system.
\subsection{Point Source Artifact patterns for UBP and FDM}
\begin{figure}[!t]
    \centering
    \includegraphics[width=0.8\columnwidth]{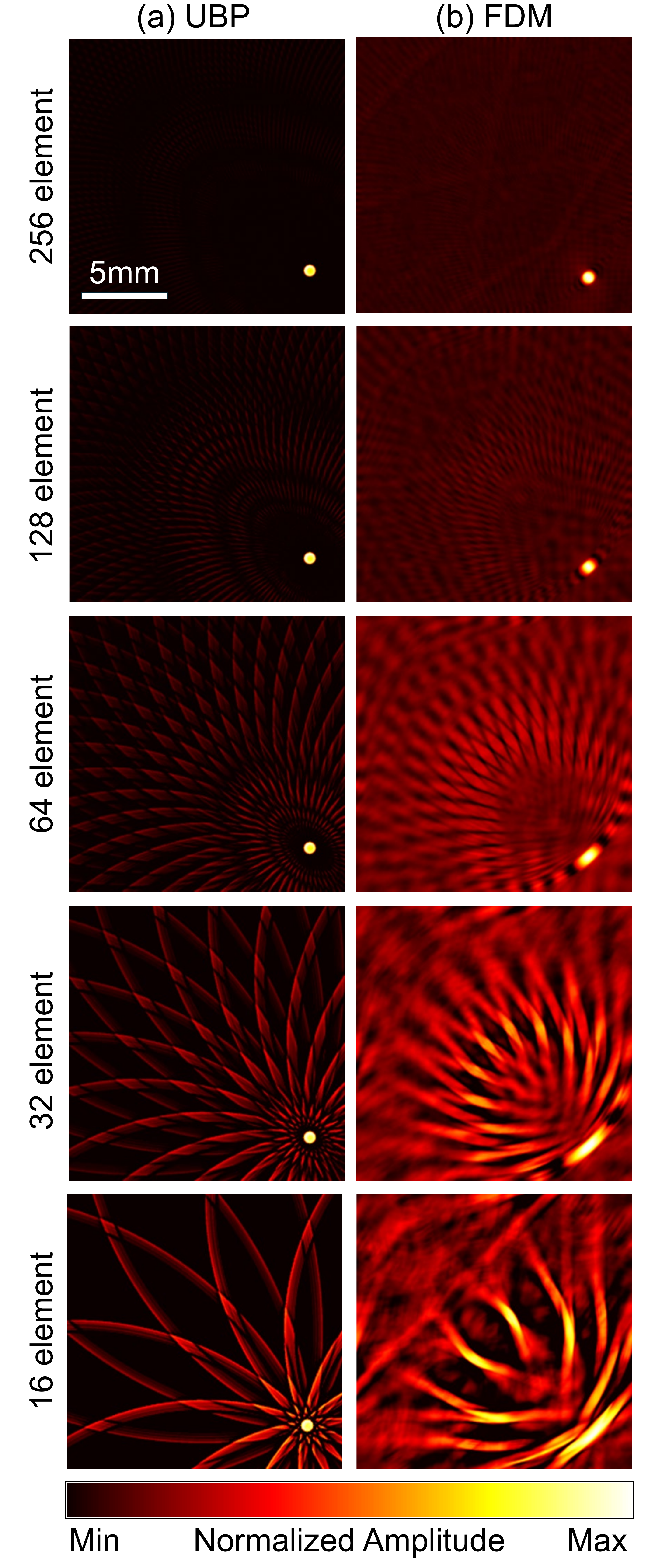}
    \caption{Point spread functions (PSFs) under sparse sampling. (a) - (b) FDM and UBP reconstructions of a point target using 256, 128, 64, 32, and 16 transducer elements.}
    \label{fig:psf}
\end{figure}

To visually represent this artifact pattern, Fig.~\ref{fig:psf} illustrates a simulation study result that shows reconstructed images of a point source obtained with full-ring array with different sensor numbers. The ring has a radius of 4 cm and 256 uniformly distributed transducers. The number of active array elements was progressively reduced from 256 to 16 to emulate increasingly sparse-view acquisition. As the number of elements decreases, both UBP and FDM algorithms suffer from severe image degradation, yet they exhibit distinct artifact morphologies. Fig.~\ref{fig:psf}(a) shows the results for the UBP algorithm, where the increased sensor sparsity interval produces sparse-view artifacts that appear as distinct, highly directional arcs around the point target. In contrast, Fig.~\ref{fig:psf}(b) shows the results for the FDM algorithm, where sparse sampling induces frequency aliasing, generating global interference ripples spreading to the entire FOV.

Therefore, UBP and FDM produce different artifact patterns due to their different reconstruction mechanisms. Specifically, UBP exhibits local directionality, whereas FDM displays global periodicity. This difference provides complementary information forms the basis for the network to remove these artifacts.

\subsection{Self-supervised Dual-domain Artifact Removal Network}

Although UBP and FDM produce distinct artifacts, simply extracting their shared features via linear methods such as Singular Value Decomposition (SVD) is insufficient. SVD assumes true signals can be isolated as shared linear components. However, for PACT, true anatomical structures and artifacts are non-linearly coupled\cite{tian2021spatial}\cite{hu2020spatiotemporal}. Thus, linear intersection cannot separate overlapping artifacts.To address this, we employ a Siamese Network to learn the complex non-linear mapping from degraded inputs to clean images. Let $Y_{UBP}$ and $Y_{FDM}$ denote the input data, which are two images reconstructed by the UBP and the FDM algorithms, respectively. The network $G_\theta$ decouples the signal by predicting artifact residuals, thereby obtaining the estimated denoised result$\hat{V}$:
\begin{align}
\hat{V}_{UBP} &= Y_{UBP} - G_\theta(Y_{UBP}) \\
\hat{V}_{FDM} &= Y_{FDM} - G_\theta(Y_{FDM})
\end{align}
To optimize $G_\theta$ without ground truth, we propose a physics-informed loss that aligns cross-domain biological structures while suppressing domain-specific artifacts. The loss function comprises two terms: cross-domain fidelity and uncertainty-weighted consistency:

\begin{equation}
\mathcal{L}_{total} = \mathcal{L}_{cross} + \lambda_{cons}\mathcal{L}_{cons}
\end{equation}
To prevent the network from simply memorizing input artifacts, we propose a cross-domain supervision strategy. Since $Y_{\text{UBP}}$ and $Y_{\text{FDM}}$ share the same anatomy, the denoised $\hat{V}_{\text{UBP}}$ should structurally and statistically match $Y_{\text{FDM}}$, and vice versa. This symmetric loss promotes bidirectional consistency between the spatial and Fourier domains, so that each domain regularizes the restoration in the other:
\begin{equation}
\mathcal{L}_{cross}
=
\lambda_{cross} \mathcal{D}(\hat{V}_{\mathrm{UBP}}, Y_{\mathrm{FDM}})
+
(1 - \lambda_{cross}) \mathcal{D}(\hat{V}_{\mathrm{FDM}}, Y_{\mathrm{UBP}})
\end{equation}
where $\mathcal{D}(\cdot,\cdot)$ denotes a point-wise reconstruction discrepancy, implemented as the Mean Square Error (MSE) loss. 
The coefficients $\lambda_{cross}$ controls the relative contributions of the two cross-domain supervision terms.

We further introduce an uncertainty-aware consistency loss to account for the spatially varying reliability of the two initial reconstructions. Since $Y_{\mathrm{UBP}}$ and $Y_{\mathrm{FDM}}$ are reconstructed from the same measurements, anatomical structures reliably recovered by both methods should exhibit small voxel-wise discrepancies, whereas large local discrepancies are more likely associated with algorithm-specific artifacts. We therefore use the cross-domain discrepancy as a proxy for reconstruction uncertainty and convert it into a reliability weight via exponential decay:

\begin{equation}
\mathcal{L}_{\mathrm{cons}}
=
\mathbb{E}_{\mathbf r}
\left[
e^{-\alpha \left|Y_{\mathrm{UBP}}(\mathbf r)-Y_{\mathrm{FDM}}(\mathbf r)\right|}
\left(
\hat{V}_{\mathrm{UBP}}(\mathbf r)
-
\hat{V}_{\mathrm{FDM}}(\mathbf r)
\right)^2
\right],
\end{equation}

where $\alpha>0$ controls the decay rate of the reliability weight. This loss enforces stronger consistency in reliable regions while suppressing the influence of inconsistent algorithm-specific artifacts.

\section{Results}
\subsection{Analysis of Reconstruction Artifacts}
\begin{figure*}[!t]
    \centering
    \includegraphics[width=\textwidth]{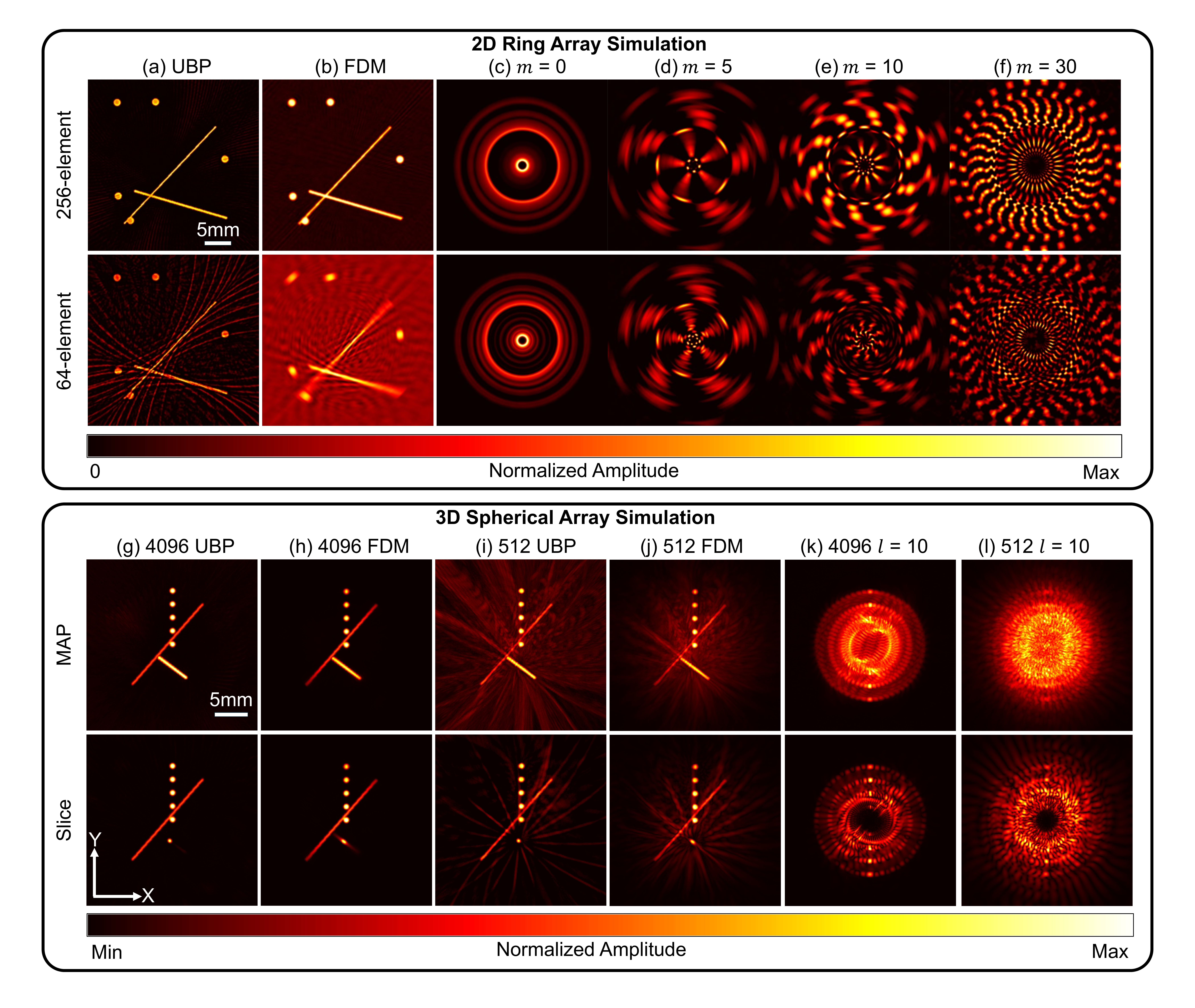}
    \caption{Artifacts of UBP and FDM. (a) - (b) 2D UBP and FDM reconstructions with 256 and 64 elements. (c) - (f) 2D angular modes ($m=0, 5, 10, 30$). (g) - (h) 3D UBP and FDM reconstructions with 4096 elements. (i) - (j) Corresponding reconstructions with 512 elements. (k) - (l) Tenth-order spherical harmonic modes with 4096 and 512 elements.}
    \label{fig:fig2}
\end{figure*}

\begin{figure*}[!t]
    \centering
    \includegraphics[width=\textwidth]{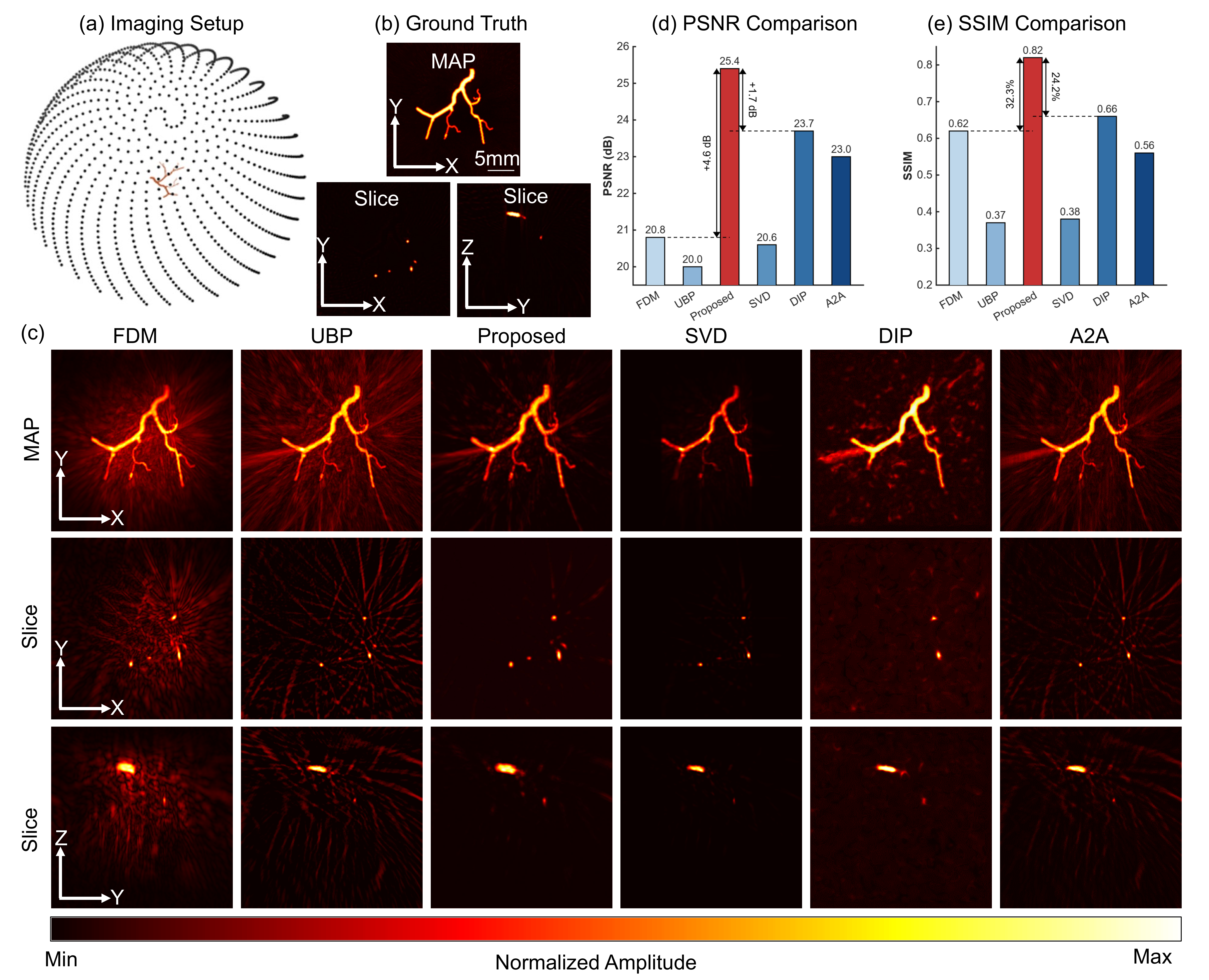}
    \caption{Image-quality assessment of the complex vascular phantom under sparse sampling. (a) Spatial distribution of phantom and detectors. (b) MAP and orthogonal slices of ground truth. (c) MAP and orthogonal slices reconstructed by FDM, UBP, proposed method, SVD, DIP, and A2A. (d) - (e) PSNR and SSIM of the reconstructed volumes.}
    \label{fig:fig3}
\end{figure*}

\begin{figure*}[!t]
    \centering
    \includegraphics[width=\textwidth]{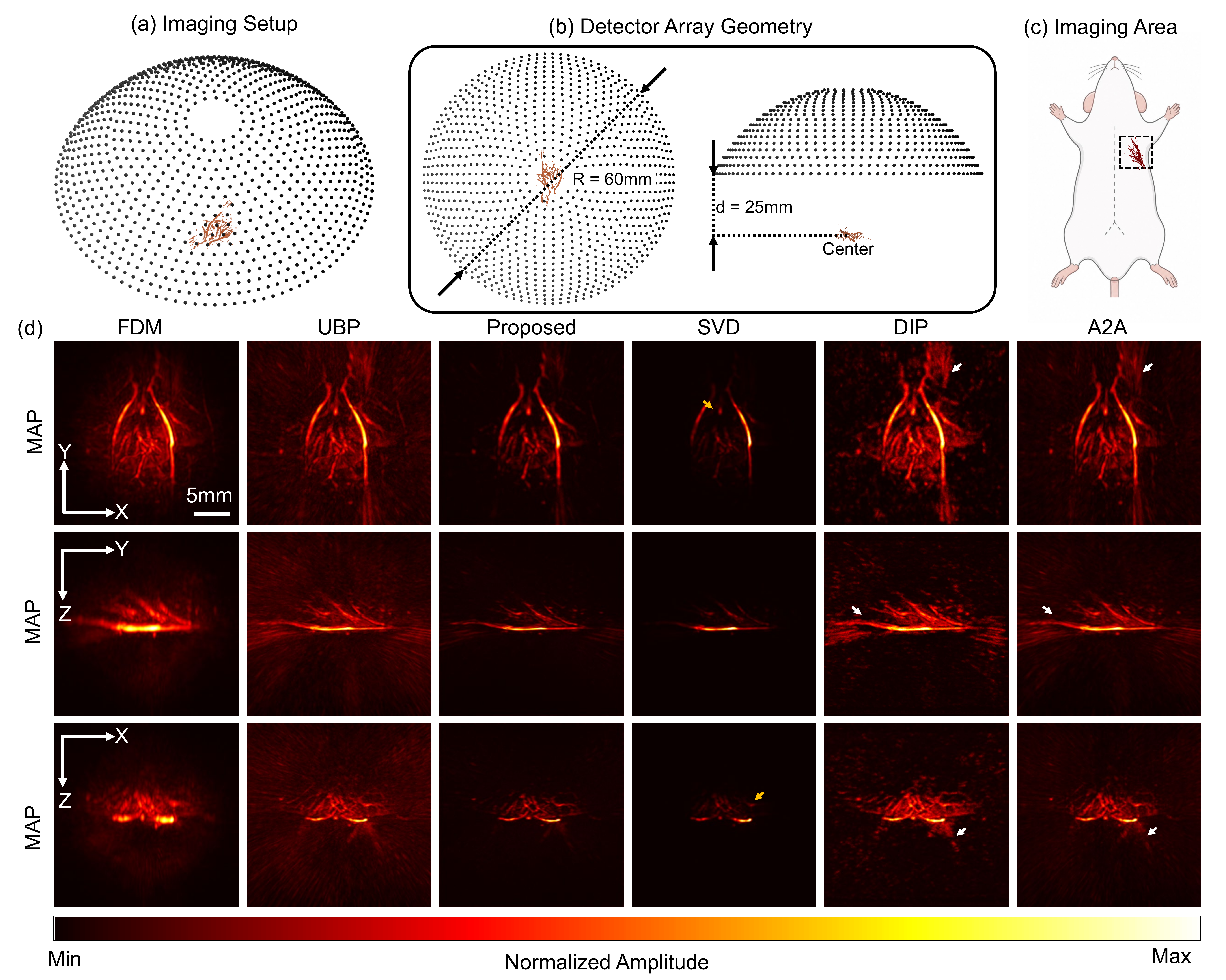}
    \caption{Evaluation on 3D \textit{in vivo} rat liver data. (a) - (c) Experimental PACT system geometry and rat liver imaging region. (d) X-, Y-, and Z-axis MAPs reconstructed by FDM, UBP, DIP, A2A, SVD, and Proposed. White arrows indicate residual artifacts, and orange arrows indicate fine anatomical details.}
    \label{fig:fig5}
\end{figure*}

\begin{figure*}[!t]
    \centering
    \includegraphics[width=\textwidth]{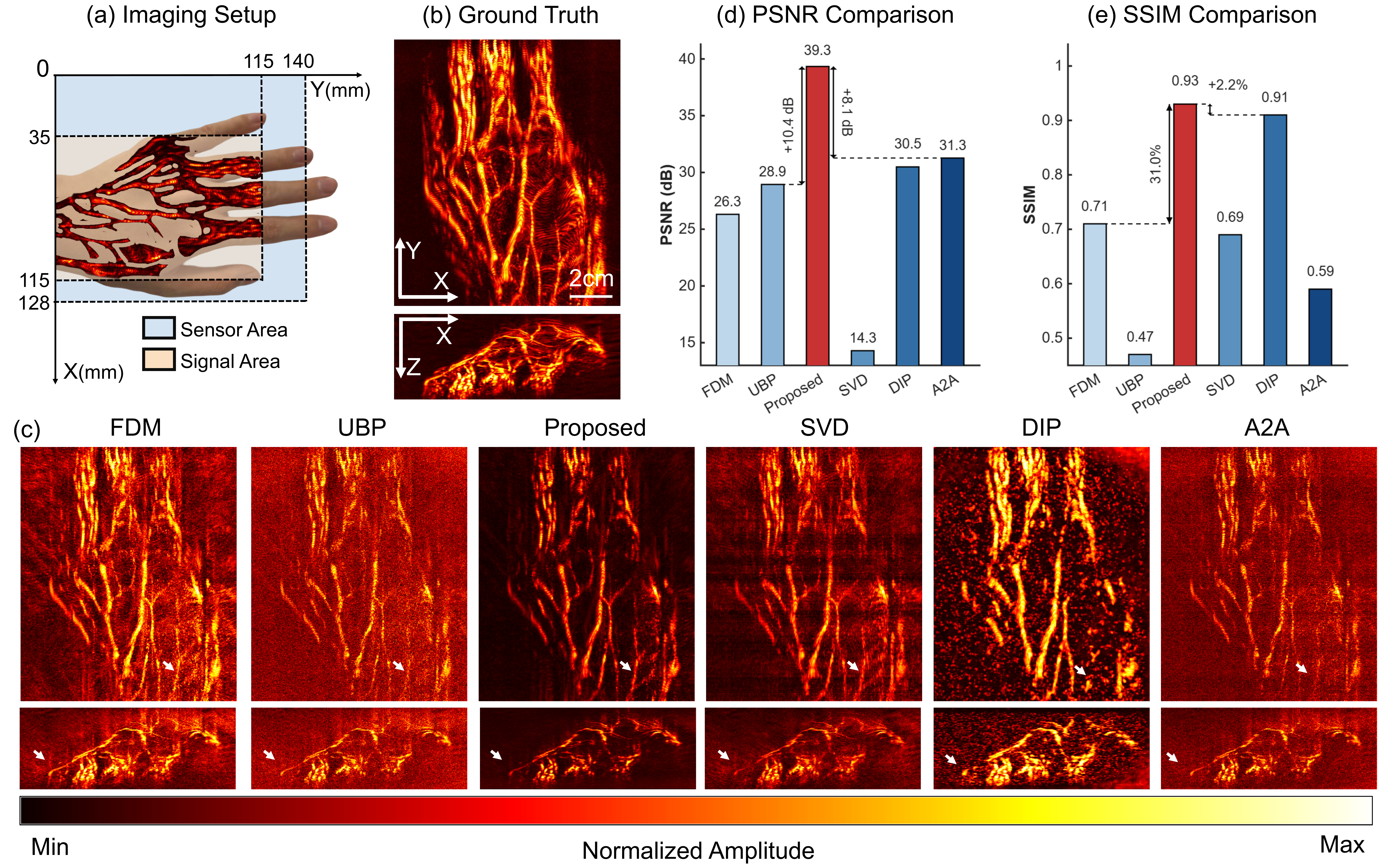}
    \caption{Image-quality assessment of the \textit{in vivo} human arm under sparse sampling. (a) Detector distribution and acquisition region. (b) Ground truth. (c) Z- and Y-axis MAPs of FDM, UBP, DIP, A2A, SVD, and Proposed. (d) - (e) PSNR and SSIM metrics. White arrows indicate preserved small blood vessels.}
    \label{fig:fig4}
\end{figure*}

Fig.~\ref{fig:fig2} compares the distinct propagation patterns of sparse-sampling artifacts produced by UBP and FDM in 2D and 3D detection geometries, providing complementary information for the proposed dual-domain reconstruction framework.

In the 2D experiment, a full-ring array with a radius of 4~cm and 256 uniformly distributed transducers was used. Sparse-view acquisition was simulated by retaining 64 transducers, corresponding to fourfold spatial downsampling. Under full sampling, both methods recovered the main target structures with weak background artifacts. After downsampling, the two methods showed distinct degradation patterns. The UBP result contained pronounced arc-shaped and directional streak artifacts around the targets, as shown in Fig.~\ref{fig:fig2}(a). This behavior arises because the continuous back-projection integral over the detection boundary in Eq.~\eqref{eq:ubp} is approximated by a sparse discrete sum. Consequently, the back-projected waves from different detector positions do not fully cancel outside the true source locations. By contrast, the FDM result exhibited widely distributed oscillatory and speckle-like artifacts, as shown in Fig.~\ref{fig:fig2}(b). The spatial displacement and shape distortion became more evident away from the center of the FOV. These observations are consistent with angular-mode undersampling and spectral aliasing during Fourier-domain inversion.

To further examine the formation of FDM artifacts, Figs.~\ref{fig:fig2}(c)~-~(f) show the spatial-domain reconstruction components obtained by retaining only the angular modes $m=0$, $5$, $10$, and $30$, respectively. According to Eq.~\eqref{eq:fft}, the frequency-domain measurements on a circular array are expanded using the angular basis $\Psi_m(\phi_s)=e^{im\phi_s}$, yielding the corresponding modal coefficients $P_m(k)$. When only a specified order $m$ is retained in Eq.~\eqref{eq:pfft_2d}, the associated single-mode object spectrum is given by
\begin{equation}
\widetilde p_m(k,\phi_k)
=
\frac{2(-i)^m\mathcal W_m(k)P_m(k)}
{\pi kH_{|m|}^{(1)}(kR_s)}
e^{im\phi_k}.
\end{equation}
After gridding and the 2D inverse Fourier transform in Eq.~\eqref{eq:pfft_recon}, this spectrum forms the corresponding spatial-domain components shown in Figs.~\ref{fig:fig2}(c)~-~(f). As $|m|$ increases, the angular oscillations become denser and more sensitive to the detector spacing. When the number of detectors is insufficient to satisfy the angular Nyquist criterion, high-order modal coefficients cannot be estimated accurately and become aliased with other discrete modes. These errors propagate across the FOV through spectral mapping, gridding, and inverse Fourier transformation, producing background fluctuations, rotational interference patterns, and position-dependent signal shifts. Because the same angular error corresponds to a larger spatial displacement at a greater radial distance, the distortion is more pronounced near the periphery of the FOV.

Similar method-specific artifacts were observed in the 3D hemispherical-array experiment. The array had a radius of 6~cm and contained 4096 transducers distributed on a Fibonacci spherical lattice. Fig.~\ref{fig:fig2}(g) and Fig.~\ref{fig:fig2}(h) show the UBP and FDM reconstructions obtained with the full array, respectively. Sparse acquisition was simulated by retaining 512 transducers, corresponding to eightfold spatial downsampling. Under this condition, UBP produced pronounced directional streak artifacts, as shown in Fig.~\ref{fig:fig2}(i), while FDM produced speckle-like interference artifacts, as shown in Fig.~\ref{fig:fig2}(j). For the 3D spherical geometry, the angular basis in Eq.~\eqref{eq:fft} is $\Psi_{lm}(\Omega_s)=Y_l^m(\theta_s,\phi_s)$, and Eq.~\eqref{eq:pfft_3d} constructs the 3D object spectrum from the spherical harmonic coefficients $P_{lm}(k)$. Fig.~\ref{fig:fig2}(k) and Fig.~\ref{fig:fig2}(l) compare the reconstruction components of the tenth-order spherical harmonic mode under full and sparse sampling. Spatial undersampling alters the modal energy distribution and causes spectral leakage among spherical harmonic modes. After 3D spectral mapping and inverse Fourier transformation, this leakage produces broadly distributed radial interference structures across the field of view.

Overall, the complementary artifact patterns of UBP and FDM arise from incomplete cancellation along sparse back-projection trajectories and angular-mode aliasing in Fourier-domain inversion, respectively, providing a physical basis for the proposed cross-domain self-supervised artifact-removal strategy.

\subsection{Numerical Phantom Evaluation}
\label{subsec:Phantom_Evaluation}

Then, we evaluated the proposed dual-domain artifact removal method against traditional image denoising techniques using a complex vascular phantom. The spatial distribution of the phantom and detectors is illustrated in Fig.~\ref{fig:fig3}(a), where 4096 transducers are uniformly arranged on a spherical surface with a $6 \text{ cm}$ radius according to a Fibonacci grid. With an array density sufficient to completely mitigate sampling artifacts, the resulting full-array image is adopted as the ground truth (Fig.~\ref{fig:fig3}(b)). To generate sparse sampling conditions, we applied an 8-fold uniform downsampling to the detector array, leaving 512 channels for reconstruction. Fig.~\ref{fig:fig3}(c) visualizes the maximum amplitude projection (MAP) along the $z$-axis alongside central slices in orthogonal directions. We compared our method with baseline FDM and UBP reconstructions, as well as two recently reported unsupervised denoising methods: Deep Image Prior (DIP)\cite{ulyanov2018deep} and Zeroshot-Artifact2Artifact (A2A)\cite{LI2025100723}. Furthermore, the SVD result is also provided, demonstrating that a simple intersection of FDM and UBP images fails to effectively suppress artifacts.

Qualitative observations of the cross-sectional slices and MAP images (Fig.~\ref{fig:fig3}(c)) highlight the limitations of conventional methods. While A2A removes some artifacts, it leaves noticeable residuals. Conversely, DIP incorrectly discards real structural details. Although SVD filters out non-shared artifacts, it aggressively removes genuine microvessels, treating them as noise. This results in severe structural fragmentation and over-smoothing. In contrast, our proposed network avoids simplistic intersection. By leveraging cross-domain non-linear supervision, it effectively eliminates complex artifacts while preserving the complete vascular morphology. These visual improvements are supported by quantitative metrics (Fig.~\ref{fig:fig3}(d) and Fig.~\ref{fig:fig3}(e)). Our method improves PSNR and SSIM by 4.6~dB and $32.3\%$, respectively, compared to the baseline FDM reconstruction. Even against the second-best method (DIP), our network achieves a gain of 1.7~dB in PSNR and a relative gain of $24.2\%$ in SSIM.

\subsection{In Vivo Animal Study Evaluation}
We further validated the proposed method on 3D in vivo imaging data of a rat liver. The data was acquired by Prof. Chulhong Kim's laboratory using a hemispherical ultrasound array consisting of 1024 elements, as illustrated in Fig.~\ref{fig:fig5}(a). The array had a radius of 60 mm and covered a solid angle of $1.15\pi$, as shown in Fig.~\ref{fig:fig5}(b). Each array element has a central frequency of 2.02 MHz and a bandwidth of 54\%. The effective FOV was $25.6 \times 25.6 \times 25.6~\text{mm}^3$, with an approximately isotropic spatial resolution of $380~\mu\text{m}$ ~\cite{https://doi.org/10.1002/advs.202202089}. The data used in this study corresponded to a rat liver imaging experiment, indicated in Fig.~\ref{fig:fig5}(c). More details about the 3D PACT system and animal experiment can be found in~\cite{https://doi.org/10.1002/advs.202202089}\cite{kim2022deep}\cite{yang2025multiplane}\cite{kim20243d}\cite{jeong2026hybrid}\cite{choi2026photoacoustic}.

For visualization, we extracted the MAPs along the $X$, $Y$, and $Z$ axes, as shown in Fig.~\ref{fig:fig5}(d). We compared the proposed method with FDM and UBP reconstructions, as well as representative DIP, A2A, and SVD methods. Qualitatively, the proposed method substantially reduced the reconstruction artifacts observed in the baseline UBP result and achieved superior artifact suppression compared with the existing denoising methods. In particular, compared with DIP and A2A, our method more effectively removed residual artifacts in the regions indicated by the white arrows. Meanwhile, compared with SVD, the proposed method better preserved structural continuity and fine anatomical details in the regions indicated by the orange arrows. These results demonstrate the robustness and generalizability of the proposed method for complex 3D experimental PACT data.

\subsection{In Vivo Human Study Evaluation}
Then, we validated the proposed method using \textit{in vivo} photoacoustic imaging data from a human hand. The PACT system was based on a custom unfocused linear array detector and performed synthetic 3D photoacoustic imaging through linear translational scanning. The detector array had 256 elements with a pitch of 0.5 mm, a central frequency of 3.5 MHz, and a -$6$ dB fractional bandwidth exceeding 80\%. Fig.~\ref{fig:fig4}(a) shows the detector aperture distribution and the photoacoustic imaging region of the hand. More details about the imaging system and the human experiment can be found in \cite{10.1117/1.JBO.29.S1.S11519}. The original data were densely sampled with a scanning step of 0.1 mm, and the corresponding reconstruction was used as the reference ground truth (Fig.~\ref{fig:fig4}(b)). To study imaging artifacts under sparse-sampling conditions, we uniformly downsampled the original data along the element direction ($x$-axis) and the scanning direction ($y$-axis), resulting in an equivalent sampling interval of 1.5 mm in both directions. 

For visualization, we extracted the maximum intensity projections (MAP) along the $Z$ and $Y$ axes (Fig.~\ref{fig:fig4}(c)). We compared our approach with FDM and UBP reconstructions, as well as representative DIP, A2A, and SVD methods. Qualitatively, our method suppressed background artifacts more effectively than all baseline and denoising methods, while better preserving the continuity of small blood vessels (white arrows). To quantify the reconstruction quality, we calculated the PSNR and SSIM metrics (Fig.~\ref{fig:fig4}(d) and Fig.~\ref{fig:fig4}(e)). Compared with the best baseline reconstruction method, our approach improved the PSNR and SSIM by 10.14~dB and $31.0\%$, respectively. Compared with the best denoising method, our approach increased the PSNR and SSIM by 8.1~dB and $2.2\%$, respectively.

\section{Ablation Studies}

Then, we conducted a systematic ablation study by comparing our method against a baseline strategy. In this baseline, the uncertainty weighting is removed, reducing the total loss to a uniform consistency loss: $\mathcal{L}_{total} = \mathcal{L}_{cross} + \lambda_{cons}(\hat{V}_{UBP} - \hat{V}_{FDM})^2$. The experimental setup, including the transducer configuration and the diversity of imaging phantoms, strictly followed the protocols detailed in Section~\ref{subsec:Phantom_Evaluation} to ensure the robustness and reproducibility of our results across varying artifact patterns and anatomical complexities.

\begin{figure}[!b]
    \centering
    \includegraphics[width=\columnwidth]{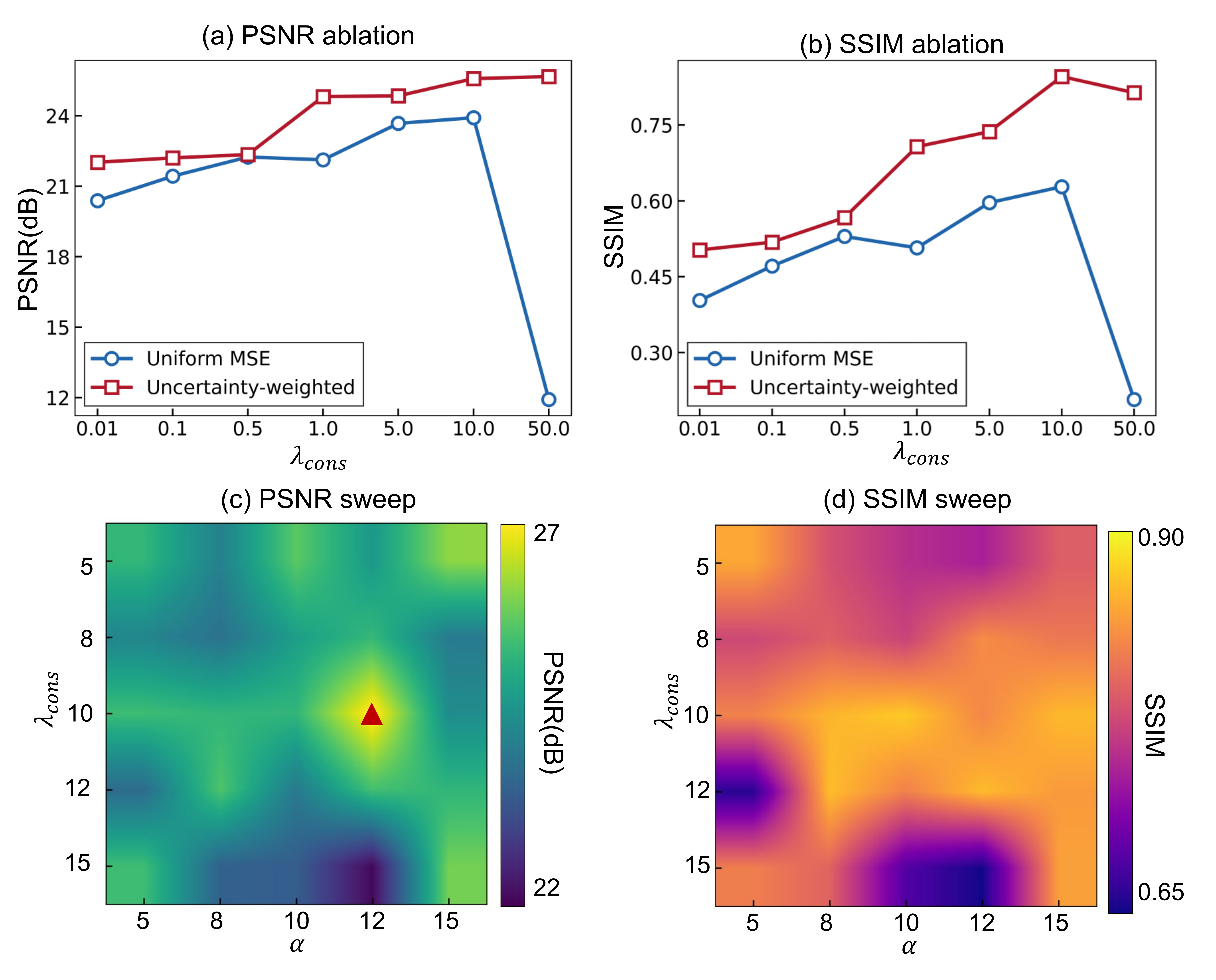}
    \caption{Ablation study and hyperparameter sensitivity analysis. (a) - (b) PSNR and SSIM performance comparison between the baseline (Uniform MSE) and the proposed uncertainty-weighted loss across varying $\lambda_{cons}$. (c) - (d) Parameter space heatmaps for $\lambda_{cons}$ and $\alpha$; the red triangle indicates the optimal configuration ($\lambda_{cons}=10, \alpha=12$).}
    \label{fig:ablation}
\end{figure}

We first performed a comprehensive sweep of $\lambda_{cons}$ within the range of $[0.01, 50]$. For each $\lambda_{cons}$, we conducted an exhaustive grid search for $\alpha$ within the same numerical range to identify the optimal performance for each value. As illustrated in Fig.~\ref{fig:ablation}(a) and Fig.~\ref{fig:ablation}(b), the proposed uncertainty-weighted mechanism consistently outperforms the traditional uniform consistency loss in terms of both PSNR and SSIM. These results indicate that our mechanism effectively mitigates domain-specific artifacts and facilitates accurate structural alignment across modalities.

To determine the optimal hyperparameter configuration for $(\lambda_{cons}, \alpha)$, we performed a refined grid search. The results (Fig.~\ref{fig:ablation}(c) and Fig.~\ref{fig:ablation}(d)) reveal that the model achieves peak PSNR when $\lambda_{cons}=10$ and $\alpha=12$. Although the SSIM exhibits minimal variation around this optimum, we adopted $\lambda_{cons}=10$ and $\alpha=12$ as our default settings due to the significant gains in PSNR and the observed structural stability.

In addition, we investigated the influence of the cross-domain weighting factor $\lambda_{cross}$. Practical PACT systems often employ hemispherical or planar detector geometries rather than a closed spherical aperture. For a hemispherical array, incomplete angular coverage causes anisotropic limited-view distortion, particularly along the missing-view directions. Consequently, increasing the contribution of the FDM branch in the cross-domain loss may transfer this directional distortion into an otherwise cleaner UBP-based estimate.

\begin{figure}[!t]
    \centering
    \includegraphics[width=\columnwidth]{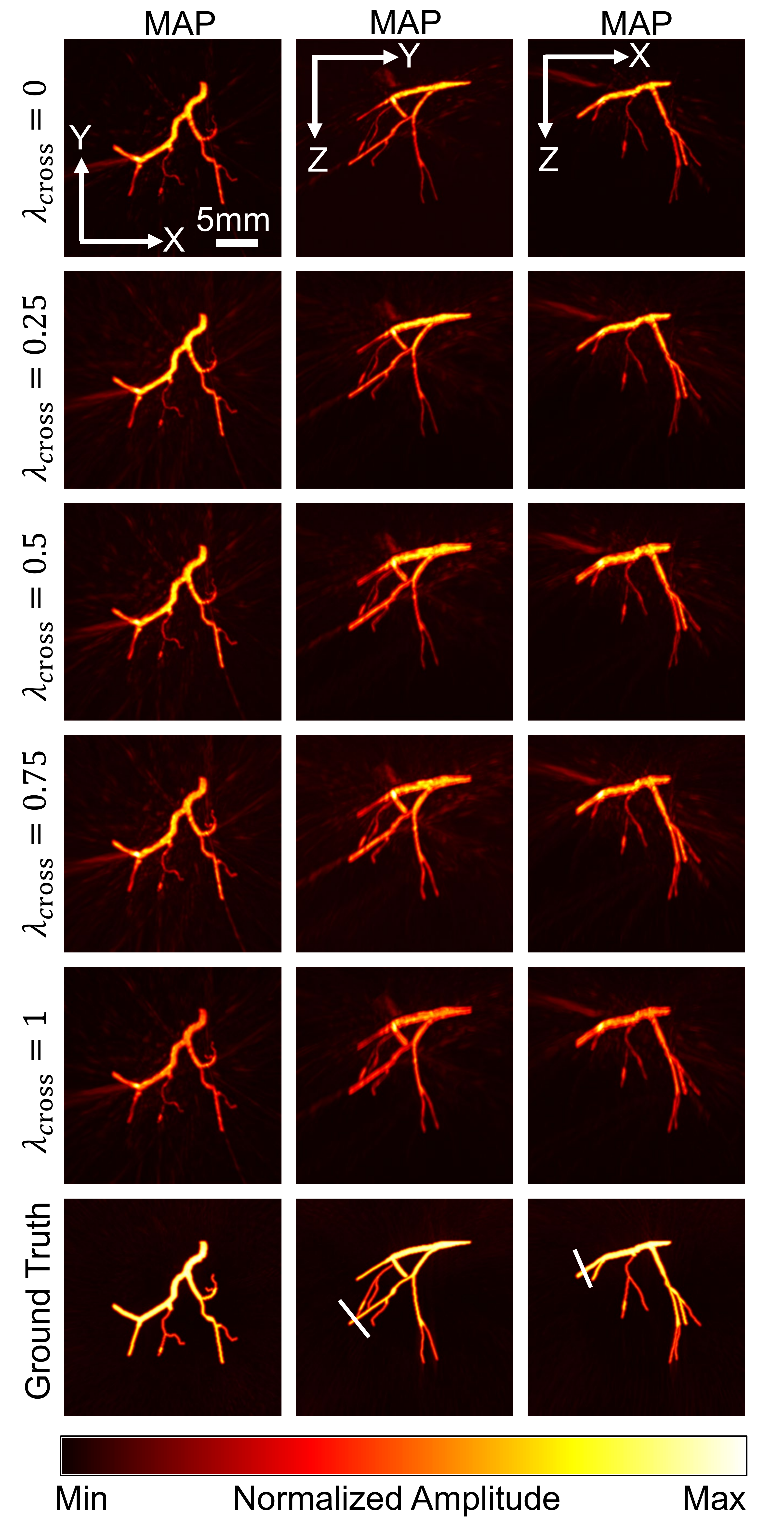}
    \caption{Maximum-amplitude-projection images obtained with different $\lambda_{cross}$ values under hemispherical-array sampling. Increasing $\lambda_{cross}$ produces progressively stronger directional distortion relative to the ground truth.}
    \label{fig:ablation_TCI1}
\end{figure}

\begin{figure}[!t]
    \centering
    \includegraphics[width=\columnwidth]{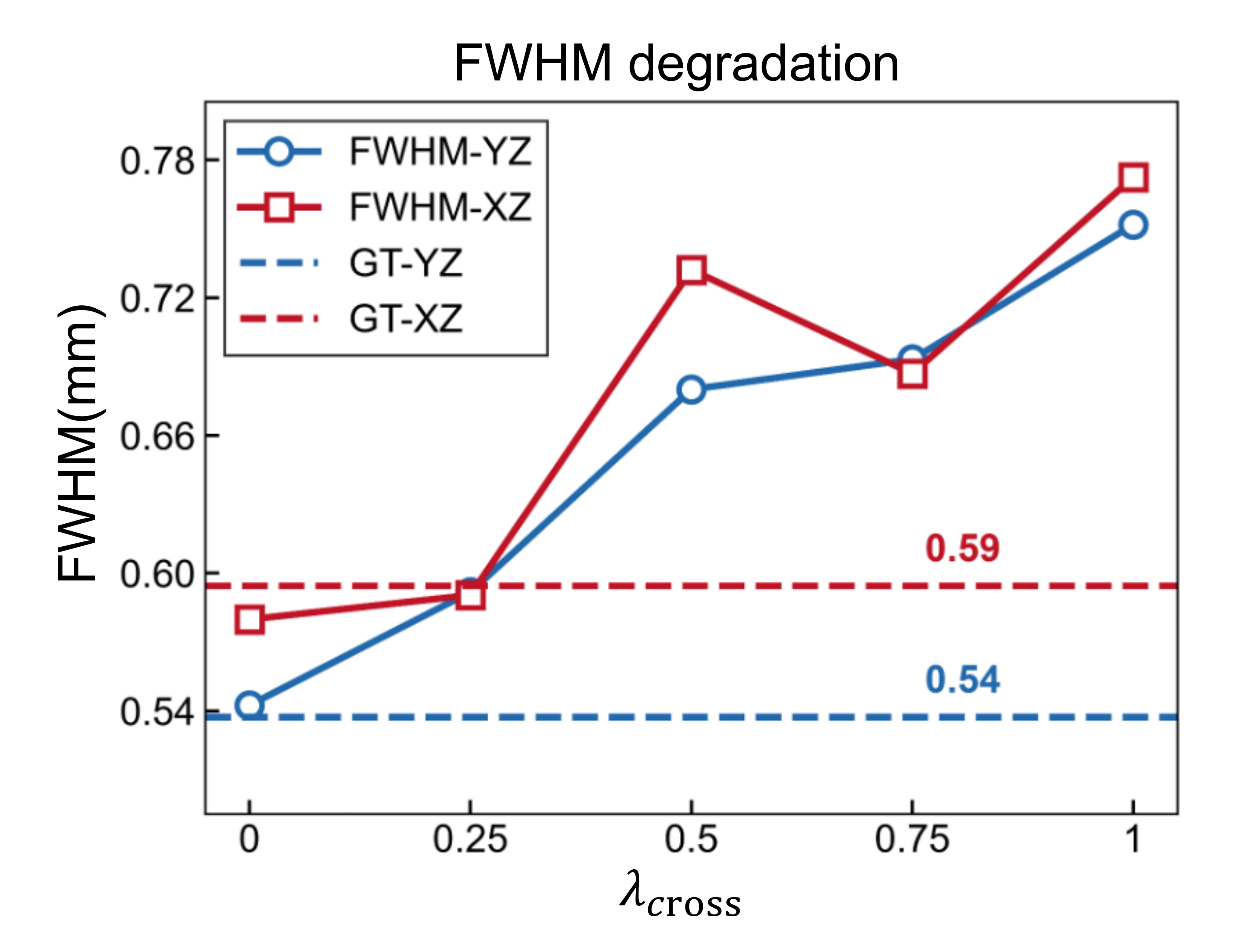}
    \caption{FWHM measurements of representative vessels for different $\lambda_{cross}$ values under hemispherical-array sampling. The ground-truth vessel widths are included as references for assessing limited-view distortion.}
    \label{fig:ablation_TCI2}
\end{figure}

To examine this effect, we performed an ablation analysis following the experimental protocol described in Section~\ref{subsec:Phantom_Evaluation}. For the hemispherical array setting, the missing-view directions are mainly along the $x$ and $y$ axes. We therefore selected representative vascular structures on the $xz$ and $yz$ planes and quantified their structural degradation using the full width at half maximum (FWHM). As shown in Fig.~\ref{fig:ablation_TCI2}, the FWHM of the selected vessels increases as $\lambda_{cross}$ increases, indicating a progressive deviation from the ground-truth FWHM. The image-domain results in Fig.~\ref{fig:ablation_TCI1} further show that the directional distortion becomes more pronounced with increasing $\lambda_{cross}$. Therefore, for both simulated and in vivo hemispherical-array experiments, we set $\lambda_{cross}=0$ to avoid propagating the limited-view distortion from the FDM reconstruction into the final output.

This degradation is specific to the missing-view geometry of the hemispherical array. In the planar-array setting, the FDM reconstruction is obtained from an analytical solution and does not introduce the same cross-constraint degradation, as demonstrated in Fig.~\ref{fig:fig4}(d). We therefore assign comparable contributions to the UBP and FDM branches and set $\lambda_{cross}=0.5$ for the planar-array experiments.

\section{Experimental Environment and Parameter Settings}
All reconstruction were performed on a platform equipped with a single NVIDIA GeForce RTX 3090 Ti GPU. We redesigned and implemented GPU-accelerated 3D FDM reconstruction algorithms for both hemispherical and planar arrays. For the simulated fully sampled hemispherical array data (original sinogram size of $4096 \times 2048$, reconstructed voxel grid of $256 \times 256 \times 256$), the reconstruction time was $5.30 \text{ s}$; whereas for the \textit{in vivo} synthetic matrix array data (original sinogram size of $300 \times 3072 \times 256$, reconstructed voxel grid of $600 \times 512 \times 160$), the reconstruction required only $0.84 \text{ s}$. This difference in efficiency primarily stems from the fact that frequency-domain reconstruction for spherical arrays must rely on spherical harmonic basis projection and non-uniform fast Fourier transform (NUFFT), these operations entail high computational complexity. In contrast, frequency-domain reconstruction for planar arrays can directly perform standard 3D-FFT on a uniform Cartesian grid, which provides a substantial computational speed advantage when handling larger-scale reconstruction tasks. The universal back-projection (UBP) reconstructions for both spherical and planar arrays were implemented based on the GAPAT architecture\cite{wang2024comprehensive}. Specifically, the UBP reconstruction time for the simulated hemispherical array data was $0.98 \text{ s}$, while the reconstruction time for the \textit{in vivo} synthetic matrix array data was $93.24 \text{ s}$.

During the training phase of the artifact removal model (covering both simulation and human experiments), the 3D reconstructed volumes were divided into 3D patches with a size of $64 \times 64 \times 64$. The hyperparameters for network training were configured as follows: the batch size was set to $2$, the initial learning rate was set to $1 \times 10^{-3}$, the number of epochs to $50$, and the learning rate decay factor ($\gamma$) to $0.6$. Under these conditions, the training times for artifact removal were $86.74 \text{ s}$ and $86.20 \text{ s}$ for the \textit{in vivo} synthetic matrix array data and the simulated hemispherical array data, respectively.

Quantitatively, the complete processing workflow comprised three sequential components: FDM reconstruction, UBP reconstruction, and network training. For the simulated hemispherical-array dataset, these components required $5.30~\mathrm{s}$, $0.98~\mathrm{s}$, and $86.20~\mathrm{s}$, respectively, yielding a cumulative runtime of $92.48~\mathrm{s}$ ($1.54~\mathrm{min}$). For the \textit{in vivo} synthetic matrix-array dataset, the corresponding runtimes were $0.84~\mathrm{s}$, $93.24~\mathrm{s}$, and $86.74~\mathrm{s}$, respectively, resulting in a total runtime of $180.82~\mathrm{s}$ ($3.01~\mathrm{min}$). Therefore, across the two evaluated datasets, the complete workflow was completed within $180.82~\mathrm{s}$ on a single NVIDIA GeForce RTX 3090 Ti GPU, quantitatively demonstrating the computational efficiency of the proposed framework.

\section{Discussion}

This study presents an efficient self-supervised artifact removal framework for PACT. The proposed method employs a lightweight Siamese Neural Network and a composite loss function integrating cross-domain fidelity and uncertainty-weighted consistency, effectively decoupling dual-domain features and filtering artifacts. Compared with single-domain reconstruction methods, either UBP or FDM, the proposed dual-domain method demonstrates prominent advantages in both PSNR and SSIM of the proposed framework. In experiments on a complex vascular phantom, compared with the baseline reconstruction method, our method improves the PSNR and SSIM by 4.6~dB and $32.3\%$, respectively. Moreover, compared with the representative existing denoising methods, our method still achieves a gain of 1.7~dB in PSNR and a relative gain of $24.2\%$ in SSIM. \textit{In vivo} rat experiments show that the proposed method effectively suppresses most sparse-sampling artifacts while preserving the integrity of vascular structures. Further human experiments demonstrate that the method can reliably retain high-frequency anatomical details, such as microvasculature. Quantitative results show that, compared with the best baseline reconstruction method, our method improves the PSNR and SSIM of \textit{in vivo} human 3D reconstructions by 10.14~dB and $31.0\%$, respectively. Compared with the best denoising method, the PSNR and SSIM are further improved by 8.1~dB and $2.2\%$, respectively.

Diven by rapid algorithmic convergence, the complete pipeline processes raw data into high-quality 3D reconstructions in under five minutes. This applies to both standard hemispherical and planar arrays using a single NVIDIA GeForce RTX 3090 Ti GPU. Furthermore, we developed GPU-accelerated frequency-domain parallel reconstruction algorithms specific to these two representative array geometries. This implementation significantly enhances the practical feasibility of applying frequency-domain inverse operators to large-scale photoacoustic datasets.

Despite these advancements, the study has certain limitations. Currently, the study has certain limitations. Currently, the frequency-domain module is only valid for circular, planar and spherical detection geometries that have complete analytical expressions, meaning it cannot be directly applied to arbitrary or irregular spatial detector geometries. However, given that photoacoustic imaging systems generally utilize planar, ring, or hemispherical transducer arrays, this framework retains substantial practical value for in vivo research and clinical translation. Notably, other tomographic modalities, such as X-ray computed tomography, also rely on dual-domain reconstruction mechanisms (e.g., filtered back-projection in the spatial domain and direct Fourier transform in the frequency domain). Exploring the generalization potential of this dual-domain self-supervised framework across broader multimodal tomographic imaging applications will be an important direction for future research.
\section*{Acknowledgment}
This research was supported by the National High-Level Hospital Clinical Research Funding (No.~2025-PUMCH-C-052), the Beijing Natural Science Foundation (No.~7232177), the National Key R\&D Program of China (No.~2023YFC2411700), and the Basic Science Research Program through the National Research Foundation of Korea (NRF), funded by the Ministry of Education (No.~2020R1A6A1A03047902). This work was also supported by the Biomedical Computing Platform of the National Biomedical Imaging Center at Peking University.
\bibliographystyle{IEEEtran}  
\bibliography{references}  


\end{document}